\documentclass[sigconf]{acmart}
\renewcommand\footnotetextcopyrightpermission[1]{}

\usepackage{amsmath}
\usepackage{dsfont}
\usepackage{algorithmicx}
\usepackage[ruled,commentsnumbered,boxed]{algorithm2e}
\usepackage{multirow}
\usepackage{graphicx}
\usepackage{subcaption}
\usepackage{xcolor} 
\usepackage{adjustbox}

\AtBeginDocument{%
  }

\acmConference[CIKM'24]{ACM Conference}{Oct 2024}{Boise, Idaho, USA}

\begin{document}

\title{Query-Enhanced Adaptive Semantic Path Reasoning for Inductive Knowledge Graph Completion}

\author{Kai Sun}
\affiliation{%
  \institution{Beijing University of Technology}
  \city{Beijing}
  \country{China}}
\email{sunkai@emails.bjut.edu.cn}

\author{Jiapu Wang}
\authornote{Both authors contributed equally to this research.}
\affiliation{%
  \institution{Beijing University of Technology}
  \city{Beijing}
  \country{China}}
\email{jpwang@emails.bjut.edu.cn}

\author{Huajie Jiang}
\affiliation{%
  \institution{Beijing University of Technology}
  \city{Beijing}
  \country{China}}
\email{jianghj@bjut.edu.cn}

\author{Yongli Hu}
\affiliation{%
 \institution{Beijing University of Technology}
  \city{Beijing}
  \country{China}}
\email{huyongli@bjut.edu.cn}

\author{Baocai Yin}
\affiliation{%
  \institution{Beijing University of Technology}
  \city{Beijing}
  \country{China}}
\email{ybc@bjut.edu.cn}

%
\renewcommand{\shortauthors}{Sun, et al.}

\begin{abstract}
Conventional Knowledge graph completion (KGC) methods aim to infer missing information in incomplete Knowledge Graphs (KGs) by leveraging existing information, which struggle to perform effectively in scenarios involving emerging entities.
Inductive KGC methods can handle the emerging entities and relations in KGs, offering greater dynamic adaptability. While existing inductive KGC methods have achieved some success, they also face challenges, such as susceptibility to noisy structural information during reasoning and difficulty in capturing long-range dependencies in reasoning paths. To address these challenges, this paper proposes the \textbf{Q}uery-Enhanced \textbf{A}daptive \textbf{S}emantic \textbf{P}ath \textbf{R}easoning (QASPR) framework, which simultaneously captures both the structural and semantic information of KGs to enhance the inductive KGC task. Specifically, the proposed QASPR employs a query-dependent masking module to adaptively mask noisy structural information while retaining important information closely related to the targets. Additionally, QASPR introduces a global semantic scoring module that evaluates both the individual contributions and the collective impact of nodes along the reasoning path within KGs. 
The experimental results demonstrate that QASPR achieves state-of-the-art performance.

\end{abstract}



\maketitle
\section{Introduction}
Knowledge Graphs (KGs) are structured graph networks where entities are represented as nodes and relations as edges. Currently, KGs are widely used in fields such as recommendation systems \cite{wei2022contrastive} and natural language processing \cite{wang2023survey}. However, due to the continuous emergence of new knowledge, the issue of incompleteness in knowledge graphs has garnered increasing attention. To address this, the task of Knowledge Graph Completion (KGC) \cite{wang2023multi} has emerged, aiming to leverage existing knowledge to infer missing facts and enhance completeness. Although traditional KGC methods \cite{10115028, ho2018rule, wang2022multi} have made some progress, they only address entities already present in KGs and struggle to perform effectively in scenarios involving emerging entities. To tackle these issues, inductive KGC methods \cite{teru2020inductive, chen2021topology, zhu2021neural, zhang2021knowledge} have been developed. 

\begin{figure}[t]
\centering
\includegraphics[scale=0.55]{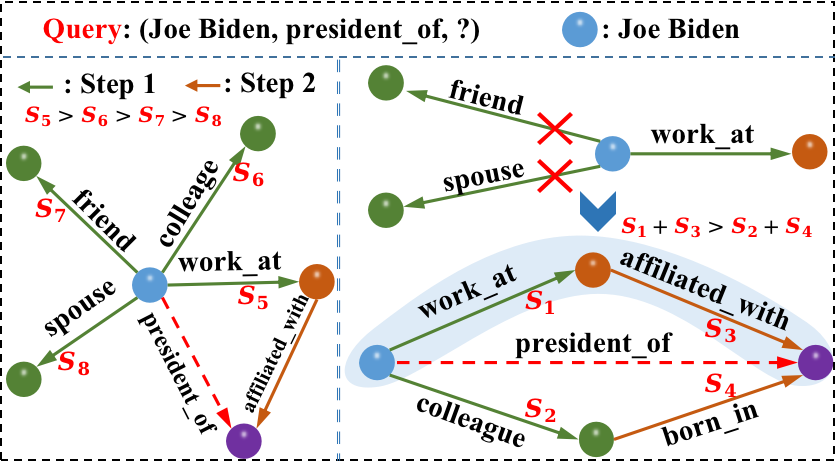}
\vspace{-10pt}
\caption{The motivation of the conventional inductive KGC (left) and the proposed QASPR (right). Conventional inductive KGCs traverse nodes directly over KGs. QASPR sequentially masks the noise structure and
captures the long-range semantic dependencies over the whole reasoning path.}
\label{path}
\end{figure}

Inductive KGC typically relies on techniques such as Graph Neural Networks (GNNs) \cite{wang2022exploring, zhu2023net} and self-supervised learning \cite{9864100} to learn general representations from the graph structure and handle the introduction of new entities. Despite the progress made by current methods, they still face numerous challenges. On the one hand, as shown in Figure \ref{path}, previous research has often overlooked the significant challenges posed by the highly complex structures of KGs. The complexity of KGs, with their numerous nodes and edges representing diverse and intricate relation types, can introduce noise during the reasoning process and hinder the effective utilization of structural characteristics \cite{wang2024made}. Thus, effectively removing the noise from irrelevant structural information within KGs is crucial.

\begin{figure*}[t]
\centering
\includegraphics[scale=0.65]{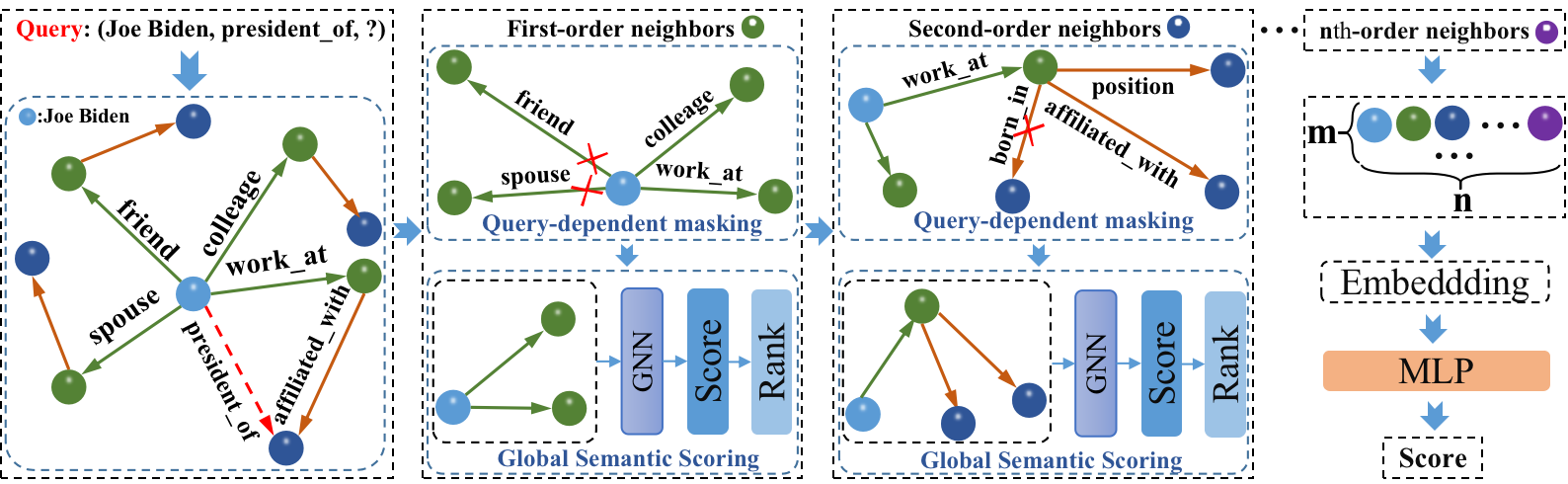}
\caption{The whole framework of QASPR.
Specifically, Query-dependent masking module adaptively masks the noisy structural information, thereby retaining effective structure; Global semantic scoring module captures the long-range dependencies by evaluating the semantics of the whole reasoning path; MLP is utilized to compute the scores for the candidates. \textbf{m} indicates the number of reasoning paths and \textbf{n} denotes number of nodes in every reasoning path.}
\label{overview}
\end{figure*}

On the other hand, prior research has typically focused only on the immediately current node in the reasoning path, overlooking the critical long-range semantic dependencies among entities throughout the entire reasoning path. 
This oversight can lead to an incomplete understanding of the semantic context, thereby reducing the effectiveness of reasoning tasks. Therefore, it is essential to develop a mechanism that captures the full scope of semantic dependencies, considering the cumulative influence of all nodes in the reasoning path.


To address these challenges, this paper proposes the \textbf{Q}uery-Enhanced \textbf{A}daptive \textbf{S}emantic \textbf{P}ath \textbf{R}easoning (QASPR) framework, which simultaneously captures both structural and semantic information of KGs to enhance the inductive KGC task. Specifically, the proposed QASPR employs a query-dependent masking module to adaptively mask the noisy edges, retaining edges closely related to the targets. Additionally, QASPR introduces a global semantic scoring module that evaluates both the individual contributions and the collective impact of nodes along the reasoning path within KGs. This module can accurately identify critical semantic paths for effective reasoning, capturing long-range semantic dependencies in the reasoning path.



To summarize, the main contributions of this paper are as follows: (1) This paper proposes a novel Query-Enhanced Adaptive Semantic Path Reasoning (QASPR) framework to capture both structural and semantic information of KGs, enhancing the inductive KGC task; (2) This paper designs a query-dependent masking module, which adaptively masks the noisy structures and retains the relevant structures within KGs in a data-driven manner; (3) This paper introduces an innovative path global semantic scoring module to capture the long-range semantic
dependencies of the reasoning path within KGs; 


\section{Preliminary}
\textbf{Inductive Knowledge Graph Completion}. A knowledge graph, denoted as $\mathcal{G} = (\mathcal{V}, \mathcal{E}, \mathcal{R})$, consists of finite sets of facts (edges) $\mathcal{E}$, entities (nodes) $\mathcal{V}$, and relations  $\mathcal{R}$. Each fact is represented as a triplet $(s, r, o) \in \mathcal{V} \times \mathcal{R} \times \mathcal{V}$, indicating a relation $r$ from the head entity $s$ to the tail entity $o$. KGC typically predicts the missing information through the known knowledge. Specifically, for a query $(s, r_q, ?)$, the goal is to find the set of answers $\mathcal{V}{(s,r_q,?)}$ such that for all $o \in \mathcal{V}{(s,r_q,?)}$, the triplet $(s, r_q, o)$ holds true. Based on KGC, inductive KGC is used to predict missing facts for emerging entities that never appear in KGs. 

\textbf{Logical Rules}. Logical rules $\rho$ \cite{sadeghian2019drum, wang2024large} define the relation between two entities $s$ and $o$:
\begin{equation}
\label{logical_rule}
\begin{aligned}
    \rho:\wedge_{i=1}^{l-1}r^{*}(s,o_i) \Rightarrow r_l(s,o),
\end{aligned}
\end{equation}
where the right-hand side denotes the rule head with relation $r$ that can be induced by ($\Rightarrow$) the left-hand rule body. The rule body is represented by the conjunction ($\wedge$) of a series of body relations $r^* \in \{r_1,...,r_{l-1}\}$. The logical rule is referred to as the single rule when $r^*= r_1$. 


\section{Method}
In this section, we introduce the Query-Enhanced Adaptive Semantic Path Reasoning (QASPR) framework for inductive KGC. QASPR mainly comprises two modules, including the query-dependent masking and global semantic scoring modules. Specifically, the query-dependent masking module employs the Bernoulli distribution \cite{zhu2021graph} to mask the noisy structures, thereby obtaining effective structural characteristics. The global semantic scoring module evaluates the semantics of the whole reasoning path to capture long-range semantic dependencies, thereby improving the accuracy of reasoning tasks. The framework is illustrated in Figure \ref{overview}.

\subsection{Query-Dependent Masking}
The query-dependent masking module utilizes the Bernoulli distribution to adaptively mask the noisy structures irrelevant to the query, thereby retaining critical structural information. Specifically, this module first extracts single rules and calculates their confidence to evaluate the relevance between relations. Subsequently, it applies a normalization strategy to convert this confidence into probability. Finally, the Bernoulli distribution is employed to filter out the irrelevant relations. 

Firstly, following logical notations, we denote potential relevance between relation $r$ and relation $r_q$ using a single rule $r \Rightarrow r_q$. To quantify the degree of such relevance, we define the confidence of a single rule as follows:
\vspace{-3pt}
\begin{equation}
\label{confidence}
\begin{aligned}
    \mathcal{C}(r \Rightarrow r_q) = \frac {\sum_{t \in \mathcal{E}}\mathds{1}(r \in E_{r}(t) \land r_q \in E_{r}(t))}{\sum_{t \in \mathcal{E}} \mathds{1} (r \in E_{r}(t))},\nonumber
\end{aligned}
\end{equation}
where the function $\mathds{1}(x)$ equals 1 when $x$ is true and 0 otherwise, $E_{r}$ denotes the extracted relations from the triplets. 
$\mathcal{C}(r \Rightarrow r_q)$ quantifies the relevance between $r$ and $r_q$; a higher value indicates stronger relevance. 

Next, we calculate the probability based on their confidence. The probability can be obtained through a normalization step that transforms the confidence into probability:
\vspace{-5pt}
\begin{equation}
\label{relevance_prob}
\begin{aligned}
    p_{rr_q}^{(l)} = \min(\frac{\mathcal{C}_{max}^{(l)}-\mathcal{C}(r \Rightarrow r_q)}{\mathcal{C}_{max}^{(l)}-\mathcal{C}_{avg}^{(l)}} \cdot p_{e},\ p_{\tau}),\nonumber
\end{aligned}
\end{equation}
where $p_{rr_q}^{(l)}$ denotes the relevance between $r$ and $r_q$, which reflects the importance of the relation $r$, probability multiplier $p_{e}$\cite{zhu2021graph} is a hyper-parameter controlling the overall probability, $\mathcal{C}_{max}^{(l)}$ and $\mathcal{C}_{avg}^{(l)}$ represent the maximum and average values within the confidence set $\mathcal{C}^{(l)}$, and $p_{\tau}$ is a cut-off probability used to truncate probability value, as extremely low probabilities can lead to the loss of important relations.

Finally, we obtain a modified subset $\widetilde{\mathcal{R}}^{(l)}$ from the candidate relations $\hat{\mathcal{R}}^{(l)}$ with certain probabilities at the $l$-th hop:

\begin{equation}
\label{bernoulli_sample}
\begin{aligned}
  \widetilde{\mathcal{R}}^{(l)} = \{r \mid r \in \hat{\mathcal{R}}^{(l)}, Bern(p_{rr_q}^{(l)}) = 1\}\nonumber,
\end{aligned}
\end{equation}
where $\hat{\mathcal{R}}^{(l)}$ denotes the set of $l$-th order neighboring relations of entity $s$, $Bern$ represents a Bernoulli distribution, $\widetilde{\mathcal{R}}^{(l)}$ is then used as the relation set.

After obtaining the subset $\widetilde{\mathcal{R}}^{(l)}$, we can effectively mask the noisy edges, thereby providing a clean KG for the subsequent reasoning process.

\subsection{Global Semantic Scoring}
 
Global semantic scoring module aims to capture long-range semantic dependencies by calculating the semantic scores along the whole reasoning path. Specifically, as the reasoning process progresses, the module continuously integrates the `current node' and transitions the previous `current node' into `historical node'. This dynamic update mechanism not only ensures the accurate transmission of semantic information during the reasoning process but also enhances the understanding and handling of complex paths.

During the \(L\)-step reasoning process, every step of the reasoning path with length \(l\) ($0<l\leq L$) treats the node reached at the step as the current node and computes its score:
\vspace{-3pt}
\begin{equation}
\begin{split}
    S_{cur}^l=\left\{
                \begin{array}{ll}
                   0, &l=0\\
                  \mathbf{W}^{T}\mathbf{Cur}(l), &1 \leq l \leq L ,\\
                \end{array}
              \right.\nonumber
\end{split}
\end{equation}
where $\mathbf{W}$ indicates the learnable parameter, $\mathbf{Cur}(l)$ is the embedding of the `current node' with the length $l$. Furthermore, the reasoning path with length $(l-1)$ $(1\leq l\leq L)$ is treated as historical nodes, and the scores can be computed as follows:
\begin{equation}
\begin{split}
    S(P_{0 \rightarrow(l-1)})=\left\{
                \begin{array}{ll}
                   0, &l=1\\
                  \sum_{i\in\{1,\ l-1\}} S_{cur}^{i}, &2 \leq l \leq L ,\\
                \end{array}
              \right.\nonumber
\end{split}
\end{equation}
where $ S_{cur}^{l-1}$ represents the score of the historical node with length $(l-1)$, and $S(P_{0 \rightarrow(l-1)})$ denotes the whole score of all historical nodes.

The global semantic scoring module treats all nodes as integral components of the reasoning path. Consequently, the path score is based on the relevance of all nodes along the reasoning path, providing a more thorough assessment compared to methods that focus solely on the current node. The detail is shown as follows:
\vspace{-3pt}
\begin{equation}
\label{scoring_mechanism}
S(P_{0 \rightarrow l}) =
S(P_{0 \rightarrow(l-1)})+ S_{cur}^l,\nonumber
\end{equation}
where $S(P_{0 \rightarrow l})$ indicates the score of the whole reasoning path with length $l$. 

\subsection{Entity Embedding} 
Entity embedding module aims to leverage the existing information in KGs to generate embeddings for emerging entities. Specifically, we first employ the global semantic scoring module to obtain the scores of various reasoning paths. Subsequently, we utilize the greedy algorithm to select the \textit{Top}-$k$ paths with the highest scores. Finally, we perform entity embedding for the emerging entities through an iterative learning strategy based on these paths.


Previous research mainly employs path-based approaches \cite{sadeghian2019drum, zhu2021neural} to obtain embeddings for emerging entities by analyzing the paths between a pair of entities within a KG \cite{wang2024large}. From a representation learning perspective, they aim to learn a representation $\mathbf{h}_{o}$ to predict the triplet $(s, r_q, o)$ based on the set of reasoning paths $\mathcal{P}$ from entity $s$ to entity $o$:
\vspace{-3pt}
\begin{equation}
\label{triple_embd}
\begin{aligned}
    \mathbf{h}_{o} = \sum_{P \in \mathcal{P}}\sum _{(u,r,v)\in P} \mathbf{h}_{(u,r,v)} = \sum_{P \in \mathcal{P}}\sum _{(u,r,v)\in P} \mathbf{W}_{r_q}^{T}[\mathbf{h}_{r_q};\mathbf{h}_{r}],
\end{aligned}
\end{equation}
where $[\cdot;\cdot]$ is the concatenation operator, $\mathbf{h}_{(u,r,v)}$ is the representation of triplet $(u, r, v)$  based on the query relation $r_q$, $\mathbf{W}_{r_q}$ is a learnable parameter, $\mathbf{h}_{r_q}$ denotes the representation of the relation $r_q$, and $\mathbf{h}_{r}$ denotes the representation of the relation $r$.

It is worth noting that there are multiple paths to reach the target entity within KGs. 
Therefore, we employ a greedy algorithm to select the \textit{Top}-$k$ paths with the highest scores:

\begin{equation}
\label{top_k_nodes}
\begin{aligned}
  \hat{\mathcal{V}}^{(l-1)} = Top\_k (\bigcup_{P \in \mathcal{P}_{0 \rightarrow (l-1)}} S(P_{0 \rightarrow (l-1)})),\nonumber
\end{aligned}
\end{equation}
where $\mathcal{P}_{0 \rightarrow (l-1)}$ denotes all the reasoning paths of the length $(l-1)$, $\hat{\mathcal{V}}^{(l-1)}$ denotes the set of the end entities inferred from $\mathcal{P}_{0 \rightarrow (l-1)}$.
Finally, due to the impacts of the $Top\_k$ strategy adopted by the greedy algorithm on entity embedding, we further revise Eq (\ref{triple_embd}) as follows:

\vspace{-10pt}
\begin{equation}
\begin{split}
\mathbf{h}_{o}^{(l)} = 
\begin{cases}
\sum\limits_{\substack{(s,r,o) \in \mathcal{E}}}\mathbf{W}_{r_q}^{T}[\mathbf{h}_{r_q};\mathbf{h}_{r}], &l=1 \\
\sum\limits_{\substack{x \in \hat{\mathcal{V}}^{(l-1)}}}\sum\limits_{(x,r,o) \in \mathcal{E}} \left( \mathbf{h}_{x}^{(l-1)}+ \mathbf{W}_{r_q}^{T}[\mathbf{h}_{r_q};\mathbf{h}_{r}] \right), &2 \leq l \leq L.\nonumber
\end{cases}
\end{split}
\label{iterative_embed_for_important_path}
\end{equation}

After the above operations, we can obtain the final embedding of the predicated entity $o$.

\subsection{Loss function} \label{subsec:loss_function}
Following the strategy \cite{lacroix2018canonical}, we train the proposed QASPR through the multi-class log-loss function $\mathcal{L}$:

\begin{equation}
\label{multi-class}
    \mathcal{L} = \sum_{(s,r_q,o) \in \mathcal{E}_{train}}( -\mathbf{W}_{s}^{T}\mathbf{h}_{o}^{(L)} + \log(\sum_{\forall {x \in \mathcal{V}}}\exp(\mathbf{W}_{s}^{T}\mathbf{h}_{x}^{(L)}))),\nonumber
\end{equation}
where $\mathbf{W}_{s} \in \mathbb{R}^{d}$ is a weight parameter, $\mathbf{h}_{o}^{(L)}$ represents the embedding of entity $o$ at step $L$, $\mathbf{h}_{x}^{(L)}$ represents the embedding of entity $x$ at step $L$, $\mathcal{E}_{train}$ denotes the set of the positive triplets $(s,r_q,o)$.

\section{Experiments}\label{Sec:Experiments}
\subsection{Experiment setup}
In this subsection, we will outline the experimental setup for the proposed QASPR framework, detailing the datasets, baselines, and parameter settings.

\textbf{Datasets.} Based on the approach in \cite{zhu2023net}, we utilize the same subsets of WN18RR and FB15k237, with each dataset having four different versions, resulting in a total of eight subsets. Each subset includes a different split of the training set and test set. For detailed statistics, please refer to \cite{zhu2023net}.

\textbf{Baselines.} The proposed QASPR is compared with several classic inductive KGC methods, including: RuleN \cite{meilicke2018fine}, NeuralLP \cite{yang2017differentiable}, DRUM \cite{sadeghian2019drum}, GraIL \cite{teru2020inductive}, NBFNet \cite{zhu2021neural}, RED-GNN \cite{zhang2021knowledge}, AdaProp \cite{zhang2023adaprop}, A*Net \cite{zhu2023net}, and MLSAA \cite{skMLSAA2024}.

\textbf{Parameter Setting.} The proposed QASPR chooses \textit{Mean Reciprocal Rank} (MRR) \cite{wang2023multi} as evaluation metric. Additionally, we adopt Adam\cite{kingma2014adam} as the optimizer. 
Furthermore, we tune the length of reasoning path $L$, the number of selected paths $K$ in entity embeddings,  probability multiplier $p_e$, cut-off probability $p_{\tau}$ in query-dependent masking module, and list the detailed information of the hyper-parameter in Table \ref{inductive_setting}.
\vspace{-5pt}
{\small
\begin{table}[h]
\centering
\caption{Hyper-parameter configurations of QASPR on both datasets.}
\vspace{-10pt}
\label{inductive_setting}
\begin{tabular}{c@{\hspace{9pt}}c@{\hspace{9pt}}c@{\hspace{9pt}}c@{\hspace{9pt}}c@{\hspace{9pt}}c@{\hspace{9pt}}c@{\hspace{9pt}}c@{\hspace{9pt}}c}
\hline
\multirow{2}{*}{Hyper-parameters} & \multicolumn{4}{c}{WN18RR} & \multicolumn{4}{c}{FB15k237} \\
\cmidrule(lr){2-5} \cmidrule(lr){6-9}
& V1 & V2 & V3 & V4 & V1 & V2 & V3 & V4 \\
\hline
$L$ & 3 & 3 & 7 & 3 & 7 & 3 & 7 & 5 \\
$K$ & 150 & 50 & 100 & 300 & 300 & 250 & 300 & 300 \\
$p_{e}$ & 0.5 & 0.3 & 0.3 & 0.6 & 0.3 & 0.7 & 0.3 & 0.4 \\
$p_{\tau}$ & 0.5 & 0.5 & 0.5 & 0.5 & 0.5 & 0.5 & 0.5 & 0.5 \\
batch size & 100 & 50 & 100 & 10 & 20 & 10 & 20 & 20 \\
\hline
\end{tabular}
\end{table}}
\subsection{Experimental Analysis} 
As shown in Table \ref{tab:results}, the experimental results validate the superiority of the proposed QASPR method compared to the current state-of-the-art baseline methods. Our method significantly improves the performance of MRR on most datasets. This indicates that QASPR can effectively mask the noisy structural information and capture long-range semantic dependencies within KGs through the query-dependent masking and global semantic scoring module, further enhancing inductive reasoning capabilities.

AdaProp\cite{zhang2023adaprop} and A*Net\cite{zhu2023net} are two important baselines, as they both facilitate the reasoning process by capturing the structural information within KGs. However, the proposed QASPR still can obtain the significant improvements on both datasets. This phenomenon demonstrates that the query-dependent masking module can effectively mask the noisy structures, thereby enhancing the inductive KGC tasks.


{\small
\begin{table}[hthp]
\centering
\caption{Performance of inductive KGC on MRR.The best score is in bold, and the second-best score is in underlined.}
\label{tab:results}
\resizebox{\columnwidth}{!}{
\begin{tabular}{lcccccccc}
\toprule
\multirow{2}{*}{Methods} & \multicolumn{4}{c}{WN18RR} & \multicolumn{4}{c}{FB15k237} \\
\cmidrule(lr){2-5} \cmidrule(lr){6-9}
& v1 & v2 & v3 & v4 & v1 & v2 & v3 & v4 \\
\midrule
RuleN & 66.8 & 64.5 & 36.8 &  62.4 & 36.3 & 43.3 & 43.9 & 42.9  \\
NeuralLP & 64.9 & 63.5 & 36.1 & 62.8 & 32.5 & 38.9 & 40.0 &  39.6 \\
DRUM & 66.6 & 64.6 & 38.0 & 62.7 & 33.3 & 39.5 & 40.2 & 41.0 \\
\cline{1-9}
GraIL & 62.7 & 62.5 & 32.3 & 55.3 & 27.9 & 27.6 & 25.1 & 22.7  \\
NBFNet & 68.4 & 65.2 & 42.5 & 60.4 & 30.7 & 36.9 & 33.1 & 30.5 \\
RED-GNN & 70.1 & 69.0 & 42.7 & 65.1 & 36.9 & 46.9 & 44.5 &  44.2\\
MLSAA & 71.6 & 70.0 & 44.8 & 65.4 & 36.8 & 45.7 & 44.2 & 43.1 \\
AdaProp & \underline{73.3} & \underline{71.5} & \underline{47.4} & \underline{66.2} & 31.0 & 47.1 & 47.1 &  45.4\\
A*Net & 72.7 & 70.4 & 44.1 & 66.1 & 45.7 & \textbf{51.0} & 47.6 & \textbf{46.6} \\
\cline{1-9}
QASPR & \textbf{79.4} & \textbf{79.8} & \textbf{52.0} & \textbf{76.4} & \textbf{48.8} & \underline{49.7} & \textbf{48.8} & \underline{46.3} \\
\bottomrule
\end{tabular}
}
\end{table}}

\subsection{Ablation Experiments}\label{subsec:ablation_studies}
In order to analyze the impact of each module, we conduct ablation experiments to validate the effectiveness of different modules. The experimental results are shown in Table \ref{Variants}. "QASPR w/o M" means removing the query-dependent masking module, and "QASPR w/o S" means removing the global semantic scoring module. The experimental results show that our proposed QASPR has a significant improvement on both datasets. We conclude that the query-dependent masking module can effectively eliminate the interference of noisy structures, and the global semantic scoring module can fully capture long-range semantic dependencies within the whole reasoning path. 
{\small
\begin{table}[h]
\centering
\caption{Ablation studies on both datasets. "QASPR w/o M" denotes the model without query-dependent masking module. "QASPR w/o S" denotes the model without the global semantic scoring.}
\vspace{-5pt}
\label{Variants}
\resizebox{\columnwidth}{!}{
\setlength{\tabcolsep}{6pt} 
\begin{tabular}{lcccccc}
\toprule
\multirow{2}{*}{Methods} & \multicolumn{3}{c}{WN18RR(V4)} & \multicolumn{3}{c}{FB15k237(V4)} \\
\cmidrule(lr){2-4} \cmidrule(lr){5-7}
& MRR & H@1 & H@10 & MRR & H@1 & H@10 \\
\midrule
QASPR w/o M & 73.3 & 62.9 & 94.4 & 45.2 & 35.2 & 63.4 \\
QASPR w/o S & 76.2 & 66.7 & 95.4 & 44.6 & 34.5 & 61.8 \\
QASPR & \textbf{76.4} & \textbf{67.2} & \textbf{95.6} & \textbf{46.3} & \textbf{36.1} & \textbf{64.5} \\
\bottomrule
\end{tabular}
}
\end{table}}

\begin{figure}[h]
\centering
\includegraphics[scale=0.3]{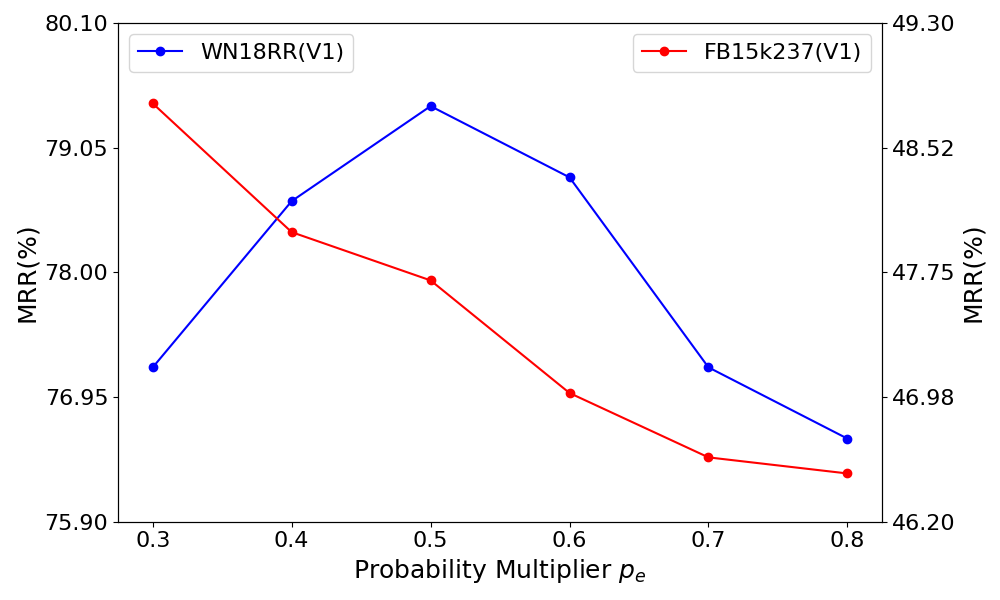}
\vspace{-10pt}
\caption{The performance with different $p_{e}$ values on WN18RR(V1) and FB15k237(V1).}
\label{pm}
\end{figure}
\subsection{Probability Multiplier $p_{e}$}
Figure \ref{pm} illustrates the trend of MRR performance for two datasets with different probability multiplier $p_e$. Overall, the MRR for the WN18RR(V1) dataset first increases and then decreases, showing a clear peak trend. In contrast, the MRR for the FB15k237(V1) dataset exhibits a continuous downward trend, with performance gradually weakening as the probability multiplier increases. This reflects the differences in how different datasets handle noise and related information. The WN18RR(V1) dataset achieves optimal performance within an intermediate range, while the FB15k237(V1) dataset shows a negative correlation with increasing probability multipliers. Therefore, selecting an optimal value of $p_e$ is crucial for maintaining model performance.

\section{Conclusion}\label{Sec:CONCLUSION}
This paper proposes a novel \textbf{Q}uery-Enhanced \textbf{A}daptive \textbf{S}emantic \textbf{P}ath \textbf{R}easoning (QASPR) framework for inductive KGC tasks. Specifically, QASPR first designs a query-dependent masking module to adaptively mask the noisy structural information, ensuring the preservation of critical information. Subsequently, QASPR develops a global semantic scoring module to capture long-range semantic dependencies in the reasoning path by evaluating the contributions of the current and historical nodes. Experimental results on two widely used datasets show the superiority of QASPR for inductive KGC tasks.

%
%
%



\balance
\bibliographystyle{unsrt}
\bibliography{sample-base}

\appendix
\end{document}